# EigenGP: Sparse Gaussian process models with data-dependent eigenfunctions

**Yuan Qi**
Departments of CS and Statistics
Purdue University
West Lafayette, IN 47906
alanqi@purdue.edu

**Bo Dai**
Department of Computer Science
Purdue University
West Lafayette, IN 47906
daibo@purdue.edu

**Yao Zhu**
Department of Computer Science
Purdue University
West Lafayette, IN 47906
zhu36@purdue.edu

## Abstract

Gaussian processes (GPs) provide a nonparametric representation of functions. However, classical GP inference suffers from high computational cost and it is difficult to design nonstationary GP priors in practice. In this paper, we propose a sparse Gaussian process model, EigenGP, based on the Karhunen-Loève (KL) expansion of a GP prior. We use the Nyström approximation to obtain data dependent eigenfunctions and select these eigenfunctions by evidence maximization. This selection reduces the number of eigenfunctions in our model and provides a nonstationary covariance function. To handle nonlinear likelihoods, we develop an efficient expectation propagation (EP) inference algorithm, and couple it with expectation maximization for eigenfunction selection. Because the eigenfunctions of a Gaussian kernel are associated with clusters of samples – including both the labeled and unlabeled – selecting relevant eigenfunctions enables EigenGP to conduct semi-supervised learning. Our experimental results demonstrate improved predictive performance of EigenGP over alternative state-of-the-art sparse GP and semisupervised learning methods for regression, classification, and semisupervised classification.

## 1 Introduction

Gaussian processes (GPs) are powerful nonparametric Bayesian models with numerous applications in statistics and machine learning. However, we face two limitations when using GPs in practice. First, it is difficult to construct *nonstationary* covariance functions for GPs because it is statistically and computationally challenging to parameterize positive definite covariance matrices as a function of the input space. In practice, while nonstationary GPs have been developed and applied to real world applications, they are often limited to low-dimensional problems, such as applications in spatial statistics (Paciorek & Schervish, 2004; Higdon et al., 1998). Second, GP inference is computationally challenging. Even for the regression case where the GP prediction formula is analytic, training the exact GP model with $N$ points demands an $O(N^3)$ computational cost for inverting the covariance matrix. To reduce the computational cost, a variety of approximate sparse GP inference approaches have been developed (Williams & Seeger, 2001; Csató & Opper, 2002; Snelson & Ghahramani, 2006; Lázaro-Gredilla et al., 2010; Williams & Barber, 1998; Qi et al., 2010) – for example, using the Nyström method to approximate covariance matrices (Williams & Seeger, 2001) or grounding the GP on a small set of (blurred) basis points (Snelson & Ghahramani, 2006; Qi et al., 2010). An elegant unifying view for various approximate sparse GP regression models is given in Quiñonero-Candela & Rasmussen (2005). Note that all these sparse GP approaches gain computational efficiency with certain approximations – possibly degenerating prediction accuracy.

In this paper, we propose a new approach, EigenGP, that addresses these two issues in a principled framework. Specifically, we project the GP prior into a space spanned by eigenfunctions, and add white noise to it to handle prediction uncertainty at infinity. The eigenfunctions depend on data input so that the covariance changes in the input space. Furthermore, based on the observed data, we select a small number of eigenfunctions by maximizing model marginal likelihood. The projection of GPs into a small eigensubspace can remove noise in function values; this is similar to what principle component analysis does in noise reduction, but we do it in a functional space for the output. Furthermore, with only a few eigenfunctions in the model, we can greatly reduce the computational cost; it is $O(NL^2)$ – rather than $O(N^3)$ – where $L$ is the number of the selected eigenfunctions. This selection also enables semi-supervised learning based on a commonly-used clustering assumption. This assumption states that if points are in the same cluster, they are likely to be of the same class. Because eigenfunctions of a Gaussian covariance function correspond to clusters of data points, we can choose clusters – based on both labeled and unlabeled data points – relevant for the predictions.

The rest of the paper is organized as follows. Section 2 describes the background of GPs. Section 3 and 4 present the EigenGP model, EP inference for EigenGP, and expectation maximization updates for sparsification. In Section 5, we discuss related works – in particular, the difference between EigenGP and the Nyström method (Williams & Seeger, 2001) and relevance vector machine (Tippings, 2000). Section 6 shows experimental results on regression, classification and semi-supervised classification, demonstrating improved predictive performance of EigenGP over state-of-the-art approaches, including support vector machines, GPs and sparse GPs.

## 2 Background of Gaussian processes

We denote $N$ independent and identically distributed samples as $\mathcal{D} = \{(\mathbf{x}_1, y_1), \ldots, (\mathbf{x}_n, y_n)\}_N$, where $\mathbf{x}_i$ is a $d$ dimensional input (i.e., explanatory variables) and $y_i$ is a scalar output (i.e., a response), which we assume is the noisy realization of a latent function $f$ at $\mathbf{x}_i$.

A Gaussian process places a prior distribution over the latent function $f$. Its projection $\mathbf{f_x}$ at $\{\mathbf{x}_i\}_{i=1}^{N}$ defines a joint Gaussian distribution: $p(\mathbf{f_x}) = \mathcal{N}(\mathbf{f}|\mathbf{m}^0, \mathbf{K})$, where, without any prior preference, the mean $\mathbf{m}^0$ are set to $\mathbf{0}$ and the covariance function $k(\mathbf{x}_i, \mathbf{x}_j) \equiv \mathbf{K}(\mathbf{x}_i, \mathbf{x}_j)$ encodes the prior notion of smoothness. A typical choice of $k$ is Gaussian covariance (or kernel)

$$k(\mathbf{x}, \mathbf{x}') = \exp\big(-\frac{||\mathbf{x}' - \mathbf{x}||^2}{2\eta^2}\big), \quad (1)$$

where $\eta$ controls the smoothness of the function. Note that this covariance function has the same value as long as $||\mathbf{x}'-\mathbf{x}||$ remains the same – regardless where $\mathbf{x}'$ and $\mathbf{x}$ are. Thus this leads to a *stationary* GP model.

For regression, we use a Gaussian likelihood function

$$p(y_i|f) = \mathcal{N}(y_i|f(\mathbf{x}_i), v_y), \quad (2)$$

where $v_y$ is the observation noise. For classification, the data likelihood has the form

$$p(y_i|f) = (1-\epsilon)\sigma(f(\mathbf{x}_i)y_i) + \epsilon\sigma(-f(\mathbf{x}_i)y_i) \quad (3)$$

where $\epsilon$ models the labeling error and $\sigma(\cdot)$ is a cumulative distribution function (cdf) of a standard Gaussian (i.e., the probit model).

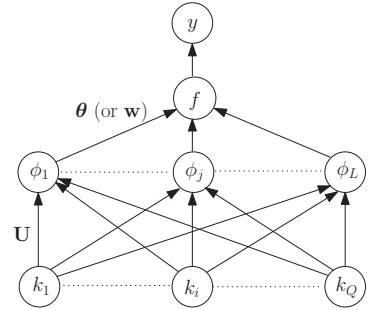

Figure 1: Deep structure of EigenGP.

Given the Gaussian process prior over $f$ and the data likelihood, the exact posterior process is

$$p(f|\mathcal{D}, \mathbf{y}) \propto GP(f|0, k) \prod_{i=1}^{N} p(y_i|f) \quad (4)$$

For the regression problem, the posterior process has an analytical form. But to make a prediction on a new sample, we need to invert a $N$ by $N$ matrix. If the training set is big, this matrix inversion will be too costly. For classification or other nonlinear problems, the computational cost is even higher because we do not have an analytical solution to the posterior process and the complexity of the process grows with the number of training samples.

## 3 Model

To obtain a nonstationary covariance function and enable fast inference, EigenGP projects the GP prior in an eigensubspace and add a white noise Gaussian process $\theta_0(\mathbf{x})$ of constant variance $w_0$ so that its prediction uncertainty does not shrink to zero at infinity. Specifically, we set the latent function $f$

$$f(\mathbf{x}) = \sum_{j=1}^{L} \theta_j \phi^j(\mathbf{x}) + \theta_0(\mathbf{x}) \quad (5)$$

where $\phi^j(\mathbf{x})$ are eigenfucntions of the GP prior. We assign a Gaussian prior over $\boldsymbol{\theta} = [\theta_1, \ldots, \theta_L]$, $\boldsymbol{\theta} \sim \mathcal{N}(\mathbf{0}, \text{diag}(\mathbf{w}))$, so that $f$ follows a GP prior with zero mean and the following covariance function

$$\tilde{k}(\mathbf{x}, \mathbf{x}') = \sum_{j=1}^{L} w_j \phi^j(\mathbf{x}) \phi^j(\mathbf{x}') + w_0 \delta_{\mathbf{x}, \mathbf{x}'} \quad (6)$$

where $\delta_{\mathbf{x}, \mathbf{x}'} = 1$ if $\mathbf{x}$ is the same as $\mathbf{x}'$ and $\delta_{\mathbf{x}, \mathbf{x}'} = 0$ otherwise. We choose $L$ in (5) to be a reasonably small number so that we can conduct efficient inference for this model as shown later.

To obtain the eigenfunctions $\phi^j(\mathbf{x})$ of a GP prior, we can use the Galekin projection to approximate them by Hermite polynomials (Marzouk & Najm, 2009). But for high dimensional problems, this approach requires a tensor product of univariate Hermite polynomials that dramatically increases the number of parameters.

To avoid this problem, we use the Nyström method (Williams & Seeger, 2001) that allows us to efficiently obtain an approximation to the eigenfunctions in a high dimensional space. Specifically, assuming basis points $\mathbf{B} = [\mathbf{b}_1, \ldots, \mathbf{b}_Q]$ ($Q \geq L$) are *i.i.d.* samples from the probability density $p(\mathbf{x})$, we can replace

$$\int k(\mathbf{x}, \mathbf{x}')\phi^j(\mathbf{x})p(\mathbf{x})\mathrm{d}\mathbf{x} = \lambda_j \phi^j(\mathbf{x}) \quad (7)$$

by its Monte Carlo approximation

$$\frac{1}{Q}\sum_{i=1}^{Q} k(\mathbf{x}, \mathbf{b}_i)\phi^j(\mathbf{b}_i) \approx \lambda_j \phi^j(\mathbf{x}) \quad (8)$$

Then, with simple derivations, we obtain the $j$-th eigenfunction $\psi^j(\mathbf{x})$ as follows

$$\phi^j(\mathbf{x}) = \frac{\sqrt{Q}}{\lambda_j^Q}\mathbf{k}(\mathbf{x})\tilde{\mathbf{u}}_j = \mathbf{k}(\mathbf{x})\mathbf{u}_j \quad (9)$$

where $\mathbf{k}(\mathbf{x}) \triangleq [k(\mathbf{x}, \mathbf{b}_1), \ldots, k(\mathbf{x}, \mathbf{b}_Q)]$, $\lambda_j^Q$ and $\tilde{\mathbf{u}}_j$ are the $j$-th eigenvalue and eigenvector of the covariance function evaluated at $\mathbf{B}$, and $\mathbf{u}_j = \frac{\sqrt{Q}}{\lambda_j^Q}\tilde{\mathbf{u}}_j$. In practice, we often select the basis points $\mathbf{B}$ as a random subset of $\mathbf{X}$. We can also first estimate $p(\mathbf{x})$ based on $\mathbf{X}$ and then sample multiple $\mathbf{B}$ from $p(\mathbf{x})$ so that we can obtain estimation uncertainty in $\phi^j(\mathbf{x})$ (i.e., $\mathbf{u}_j$).

Inserting (9) into (5), we obtain

$$f(\mathbf{x}) = \sum_{j=1}^{L}\theta_j \sum_{i=1}^{Q} u_{ij}k(\mathbf{x}, \mathbf{b}_i) + \theta_0(\mathbf{x}) \quad (10)$$

This equation reveals a two-layer structure of our model as visualized in Figure 1 (for simplicity, we do not shown $\theta_0(\mathbf{x})$ in this figure). Thus our model can be viewed as a *deep* Bayesian kernel machine. The deep structure highlights the difference between our model and the relevance vector machine (Tippings, 2000), which links the kernel function $k_i$ to $f$ directly and does not have the additional white noise $w_0$.

To learn the structure of the second layer in EigenGP, we use an empirical Bayesian technique, automatic relevance determination (ARD) (MacKay, 1992), which prunes edges (elements of $\mathbf{w}$) by maximizing model evidence (See Section 4 for more details). Since $L < N$ and $\mathbf{w}$ is sparsified, estimating the posterior process of $f$ is computationally efficient. Also, we constrain $f$ in the eigensubspace and therefore reduce noise in function values, which robustifies the model. If we have more training samples and allow a longer training time, we can increase $L$, i.e., the eigensubpsace for our model. In this fashion, EigenGP can be viewed as the method of sieves (Geman & Hwang, 1982), which has been applied to semi-nonparametric models with great success (Chen, 2007).

Note that given $\mathbf{w}$ and $\mathbf{U} = [\mathbf{u}_1, \ldots, \mathbf{u}_L]$, the prior over $f$ is nonstationary because its covariance function EigenGP in (6) varies at different regions of $\mathbf{x}$. This comes at no surprise since the eigenfunctions are tied with $p(\mathbf{x})$ in (7). This nonstationarity reflects the fact that our model is adaptive to the distribution of the explanatory variables $\mathbf{x}$.

If we use the Gaussian kernel (1) whose eigenfunctions correspond to clusters of data, we can use the cluster assumption for semi-supervised learning. More specifically, we first use both labeled and *unlabeled* data to learn clusters in the whole dataset. Assuming that most data points in the same cluster share the same label, we can then propagate the labels of points in a cluster to unlabeled data points in the same cluster. However, when labeled data points in the same cluster have different signs, our ARD sparsification allows us to automatically choose appropriate eigenfunctions to accommodating the sign change in a principled way.

### 3.1 An illustrative example

Now we give a toy example in Figure 2 to illustrate why the selection of eigenfunctions removes noise from function values and leads to easy classification. We also visualizes the nonstationarity of our model.

Given 300 samples from a mixture of four Gaussian components (See green "+" markers in the top panel of 2.a), we estimate the eigenfunctions based on a Gaussian covariance function with the kernel width $\eta = 0.2$. The 1st, 10th, 20th and 30th eigenfuntions are shown in Figure 2.(a). The eigenfunctions control the smoothness of the model. Using more eigenfunctions, we can capture more variability in function values. However, too much modeling flexibility does not help prediction accuracy; in Figure 2.(a), we can see the 10th, 20th and 30th eigenfunctions are not useful in discriminating samples from two classes, represented by red circles and black "x" markers.

Figure 2.(b) shows that our method identifies four discriminative eigenfunctions for classification; the weights of the selected eigenfunctions are shown in Figure 3.(a). The first, third and forth eigenfunctions are selected and they separates samples in the first, second and third data clusters from each other while each cluster contains samples with the same labels. What is interesting is that the second eigenfunction that covers the forth cluster is not selected, while this cluster contains labels from two classes. Instead, the ninth eigenfunction is selected and it separates samples in

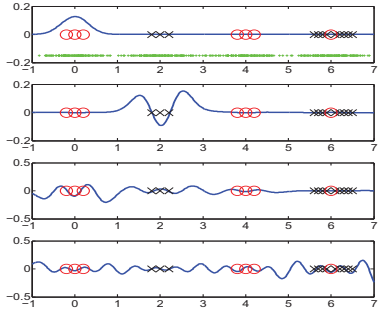

(a) 1st, 10th, 20th and 30th eigenfuntions

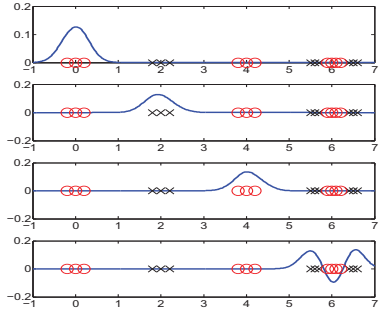

(b) Selected eigenfunctions when a cluster having samples from two classes

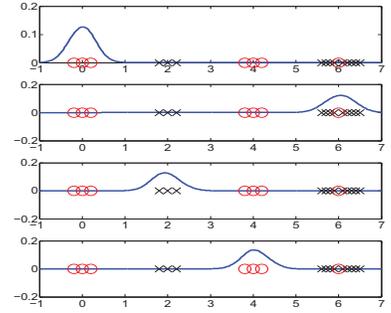

(c) Selected eigenfunctions when a cluster contains a mislabeled datapoint

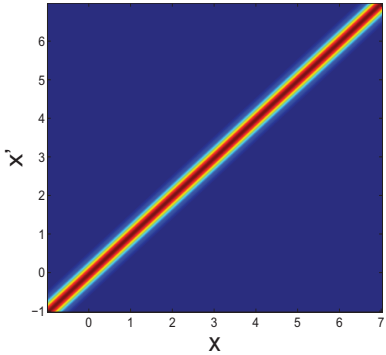

(d) Stationary Gaussian covariance function based on all the eigenfunctions

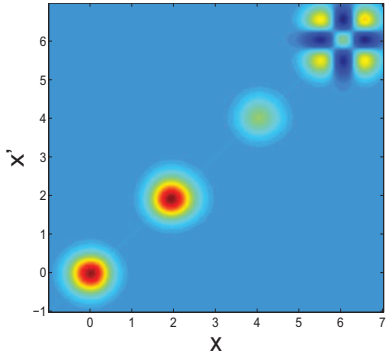

(e) Nonstationary covariance function based on the selected eigenfunctions in (b)

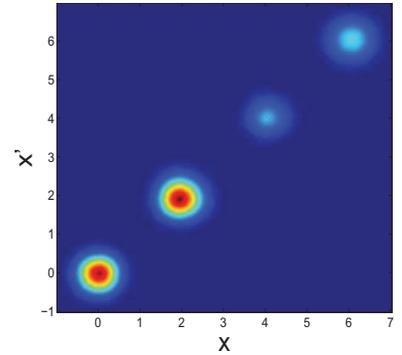

(f) Nonstationary covariance function based on the selected eigenfunctions in (c)

Figure 2: A toy example. Subfigure (a) shows the eigenfunctions of the GP model with a Gaussian covariance. (b) and (c) depict the selected eigenfunctions learned with different labeled samples. (d) shows the Gaussian covariance function evaluated at $x$ and $x'$ (both ranging from -1 to 7). It has a constant value along the diagonal direction. (e) and (f) visualize the nonstationary covariance functions corresponding to the selected eigenfunctions in (b) and (c).

the forth cluster – in other words, the samples with the same label correspond to the same sign in this eigenfunction. Therefore, based on these four selected eigenfunctions we can accurately classify the samples.

In Figure 2.(c), the samples in the forth cluster belong to one class, except a single outlier (the red circle). In this case, EigenGP automatically selects the top four eigenfunctions whose weights are shown in 3.(b). Since the second eigenfunction covers the forth cluster with the same sign, EigenGP removes the impact of the outlier, demonstrating the robustness of our model. This is similar to noise reduction in principle component analysis but in a functional space for the output.

Figures 2.(d)-(f) show the values of the covariance functions corresponding to 2, demonstrating the nonstationarity of our model.(d)-(f). As shown in 2.(d), the stationary Gaussian covariance function in (1) has

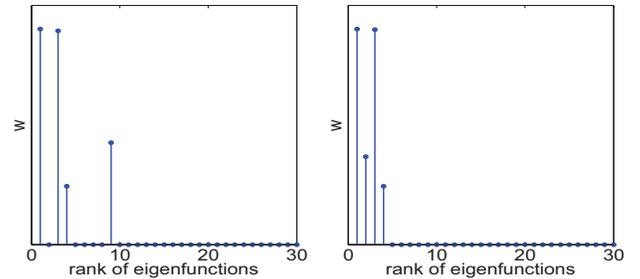

(a) Based on data in Fig. 2.(b)

(b) Based on data in Fig. 2.(c)

Figure 3: Estimated **w** for the classification of two data sets in Figure 2.(b) and (c). Only a few eigenfunctions are associated with nonzero weights.

a constant value along the diagonal direction in the image. This is because the covariance value remains the same when $\|\mathbf{x} - \mathbf{x}'\|$ is a constant regardless where the two samples $\mathbf{x}$ and $\mathbf{x}'$ are. In contrast, the covariance functions of EigenGP, defined by 6, change their values based on the values of $\mathbf{x}$ and $\mathbf{x}'$ along the diagonal direction as shown in (e) and (f). Note that the upper-right corner of (e) corresponds to the ninth eigenfunction (the bottom blue curve in (b)).

## 4  EP based evidence maximization

For Gaussian likelihoods (i.e., regression), we can use the matrix inverse lemma to efficiently compute the posterior process in (4) and the marginal likelihood. For general nonlinear likelihoods such as the probit model for classification, we can use expectation propagation (Minka, 2001) to map the nonlinear likelihood into a linear Gaussian form $\mathcal{N}(g_i|f(\mathbf{x}_i), \tau_i)$ and approximate the posterior process in (4) by

$$q(f) \propto GP(f|0, \tilde{k}) \prod_{i=1}^{N} \mathcal{N}(g_i|f(\mathbf{x}_i), \tau_i). \quad (11)$$

Let $\mathbf{g} = [g_1, \ldots, g_N]$, $\boldsymbol{\tau} = [\tau_1, \ldots, \tau_N]$, and $\Phi_\mathcal{D}$ (a $N$ by $L$ matrix) represent the values of the $L$ basis functions at the $N$ training points. Then combining (6) and (11) with equations (6.66) and (6.67) in (Bishop, 2006), we obtain the following proposition.

**Proposition 1** *The posterior process $q(f)$ defined in (11) has the mean function $m(\mathbf{x})$ and covariance function $V(\mathbf{x}, \mathbf{x}')$:*

$$m(\mathbf{x}) = \mathbf{k}(\mathbf{x})\mathbf{U}\alpha \quad (12)$$
$$V(\mathbf{x}, \mathbf{x}') = \tilde{\mathbf{k}}(\mathbf{x}, \mathbf{x}') - \mathbf{k}(\mathbf{x})\mathbf{U}\beta\mathbf{U}^\mathrm{T}\mathbf{k}(\mathbf{x}') \quad (13)$$

*where $\mathbf{W} = \mathrm{diag}(\mathbf{w})$, $\beta = \mathbf{W}\Phi_\mathcal{D}^\mathrm{T}(\tilde{\mathbf{K}} + w_0\mathbf{I} + \mathrm{diag}(\boldsymbol{\tau}))^{-1}\Phi_\mathcal{D}\mathbf{W}$, $\tilde{\mathbf{K}} = \Phi_\mathcal{D}\mathbf{W}\Phi_\mathcal{D}^\mathrm{T}$ and $\alpha = \beta\mathbf{g}$.*

From this proposition, we see the additional white noise plays a role equivalent to the Gaussian approximation of the likelihoods so that we can absorb it in the likelihoods. For the probit model, this amounts to increasing the variance of the Gaussian in the cdf.

To estimate $(\alpha, \beta)$ (equivalently, $(\mathbf{g}, \boldsymbol{\tau})$), the expectation propagation inference repeats the following three steps: message deletion, projection, and message update. In the message deletion step, we compute the partial belief $q^{\backslash i}(f; \alpha^{\backslash i}, \beta^{\backslash i})$ by removing a message $\tilde{t}_i$ from the approximate posterior $q(f; \alpha, \beta)$: $q^{\backslash i}(f; \alpha^{\backslash i}, \beta^{\backslash i}) \propto q(f; \alpha, \beta)/\tilde{t}_i$. In the projection step, we minimize the KL divergence between $\tilde{p}(f) \propto p(y_i|f)q(f; \alpha^{\backslash i}, \beta^{\backslash i})$ and the new approximate posterior $q(f; \alpha, \beta)$, such that the information from the $i$-th data point is incorporated into the model. Finally, the message $\tilde{t}_i$ is updated based on the new and old posteriors: $\tilde{t}_i \propto q(f; \alpha, \beta)/q^{\backslash i}(f; \alpha^{\backslash i}, \beta^{\backslash i})$.

### 4.1  Projection

We start with the projection step, since it is the most crucial step of EP. From (11), we see that the posterior GP of EigenGP defines an exponential family with features $\{\mathbf{f_x}, \mathbf{f_x}\mathbf{f_x}^\mathrm{T}\}$, $\mathbf{f_x} = (f(\mathbf{x}_1), \ldots, f(\mathbf{x}_{N_l}))^T$. Therefore, the minimization of the KL divergence between $\tilde{p}(f)$ and its posterior process $q(f)$ is achieved by moment matching on the mean $\mathbf{m_x}$ and the covariance $\mathbf{V_x}$ of $\mathbf{f_x}$. The moment matching equations are

$$\mathbf{m_x} = \mathbf{m_x}^{\backslash i} + \mathbf{V}^{\backslash i}(\mathbf{X}, \mathbf{x}_i)\frac{\mathrm{d}\log Z}{\mathrm{d}m^{\backslash i}(\mathbf{x}_i)} \quad (14)$$

$$\mathbf{V_x} = \mathbf{V_x}^{\backslash i} + \mathbf{V}^{\backslash i}(\mathbf{X}, \mathbf{x}_i)\frac{\mathrm{d}\log^2 Z}{\mathrm{d}m^{\backslash i}(\mathbf{x}_i)^2}\mathbf{V}^{\backslash i}(\mathbf{x}_i, \mathbf{X}) \quad (15)$$

where $Z = \int q^{\backslash i}(f)p(y_i|f)\,\mathrm{d}f$, and $m^{\backslash i}(\mathbf{x}_i)$ is the mean of $q^{\backslash i}(f(\mathbf{x}_i))$ and $\mathbf{V}^{\backslash i}(\mathbf{x}_i, \mathbf{X})$ is covariance matrix between $\mathbf{x}_i$ and $\mathbf{X}$.

Given $\mathbf{m_x}$ and $\mathbf{V_x}$, we want to compute $\alpha$ and $\beta$. To do so, first note that, from Proposition 1, it follows that

$$\mathbf{m_x} = \mathbf{K_{x,B}}\mathbf{U}\alpha \quad \mathbf{V_x} = \tilde{\mathbf{K}} - \mathbf{K_{x,B}}\mathbf{U}\beta\mathbf{U}^\mathrm{T}\mathbf{K_{x,B}^T} \quad (16)$$

where $\tilde{\mathbf{K}} = \mathbf{K_{x,B}}\mathbf{U}\mathbf{W}\mathbf{U}^\mathrm{T}\mathbf{K_{x,B}^T}$, and the $(i,j)$ element of the matrix $\mathbf{K_{x,B}}$ is $k(\mathbf{x}_i, \mathbf{b}_j)$.

Then combining (14) and (16), we obtain

$$\alpha = \alpha^{\backslash i} + \mathbf{h}\frac{\mathrm{d}\log Z}{\mathrm{d}m^{\backslash i}(\mathbf{x}_i)} \quad (17)$$

where $\mathbf{h} \triangleq \mathbf{K}^{-1}\mathbf{V}^{\backslash i}(\mathbf{X}, \mathbf{x}_i) = \mathbf{p}_i - \beta^{\backslash i}\phi_i$, $\mathbf{p}_i \triangleq \mathrm{diag}(\mathbf{w})\phi_i$ and $\phi_i \triangleq \mathbf{k}(\mathbf{x}_i)\mathbf{U}$ is precomputed before the EP inference.

Inserting (16) to (15), we obtain the update for $\beta$

$$\beta = \beta^{\backslash i} - \mathbf{h}\mathbf{h}^\mathrm{T}\frac{\mathrm{d}^2\log Z}{\mathrm{d}m^{\backslash i}(\mathbf{x}_i)^2} \quad (18)$$

The above equations define the projection step.

For the classification likelihood (3) where $\sigma(\cdot)$ is the probit function, the quantities in the projection step (17) and (18) are

$$z = \frac{m^{\backslash i}(\mathbf{x}_i)y_i}{\sqrt{v^{\backslash i}(\mathbf{x}_i, \mathbf{x}_i)}} \quad (19)$$

$$Z = \epsilon + (1 - 2\epsilon)\sigma(z) \quad (20)$$

$$\frac{\mathrm{d}\log Z}{\mathrm{d}m^{\backslash i}(\mathbf{x}_i)} = \gamma y_i \quad (21)$$

$$\frac{\mathrm{d}^2\log Z}{(\mathrm{d}m^{\backslash i}(\mathbf{x}_i))^2} = -\frac{\gamma(m^{\backslash i}(x_i)y_i + v^{\backslash i}(\mathbf{x}_i)\gamma)}{v^{\backslash i}(\mathbf{x}_i)} \quad (22)$$

where $\gamma = \frac{(1-2\epsilon)\mathcal{N}(z|0,1)}{Z\sqrt{v^{\backslash i}(\mathbf{x}_i)}}$.

### 4.2 Message update

Given the new $q(f)$, we will update the message $\tilde{t}_i(f)$ according to the ratio $q(f)/q^{\backslash i}(f)$:

$$\tilde{t}_i(f) = \frac{|\mathbf{V_x}|^{-\frac{1}{2}}\exp(-\frac{1}{2}(\mathbf{f_x} - \mathbf{m_x})\mathbf{V_x}^{-1}(\mathbf{f_x} - \mathbf{m_x}))}{|\mathbf{V_x}^{\backslash i}|^{-\frac{1}{2}}\exp(-\frac{1}{2}(\mathbf{f_x} - \mathbf{m_x}^{\backslash i})(\mathbf{V_x}^{\backslash i})^{-1}(\mathbf{f_x} - \mathbf{m_x}^{\backslash i}))}$$

$$= \mathcal{N}(f(\mathbf{x}_i)|g_i, \tau_i) \tag{23}$$

$$\tau_i \triangleq -\Big(\frac{\mathrm{d}^2 \log Z}{(\mathrm{d}m^{\backslash i}(\mathbf{x}_i))^2}\Big)^{-1} - v^{\backslash i}(\mathbf{x}_i) \tag{24}$$

$$g_i \triangleq m^{\backslash i}(\mathbf{x}_i) - \Big(\frac{\mathrm{d}^2 \log Z}{(\mathrm{d}m^{\backslash i}(\mathbf{x}_i))^2}\Big)^{-1}\frac{\mathrm{d}\log Z}{\mathrm{d}m^{\backslash i}(\mathbf{x}_i)} \tag{25}$$

where $\mathbf{v}^{\backslash i}(\mathbf{x}_i) = \boldsymbol{\phi}_i^\mathrm{T}\mathbf{h}$ and $m^{\backslash i}(\mathbf{x}_i) = \boldsymbol{\phi}_i^\mathrm{T}\boldsymbol{\alpha}^{\backslash i}$. The function $\tilde{t}_i(f)$ can be viewed as a message from the $i$-th data point to $q(f)$.

If we want to approximate the marginal likelihood of the model, we need to scale $\mathcal{N}(f(\mathbf{x}_i)|g_i, \tau_i)$ so that the message $\tilde{t}_i(f) = s_i\exp(f(\mathbf{x}_i|g_i,\tau_i)) = Zq(f)/q^{\backslash i}(f)$ preserves the local "evidence" — in other words, $\int \tilde{t}_i(f)q^{\backslash i}(f)\mathrm{d}f = \int t_i(f)q^{\backslash i}(f)\mathrm{d}f$. It is easy to show

$$\log s = \log Z - \frac{1}{2}\log(-(\log Z)''\tau_i) - \frac{((\log Z)')^2}{2(\log Z)''}$$

where $(\log Z)'$ and $(\log Z)''$ are the first- and the second- order derivatives of $\log Z$ over $m^{\backslash i}(\mathbf{x}_i)$.

### 4.3 Message deletion

To delete a message $\tilde{t}_i(f)$, we need to compute $q^{\backslash i}(f) \propto q(f)/\tilde{t}_i(f)$. Instead of computing this ratio directly, we can equivalently multiply its reciprocal with the current $q(f)$. Then we can solve the multiplication by minimizing $\mathrm{KL}(q(f)\|q^{\backslash i}(f)\tilde{t}_i(f))$ over $q^{\backslash i}(f)$. Since $q(f)$, $q^{\backslash i}(f)$, and $\tilde{t}_i(f)$ all have the form of the exponential family, the minimal value of this KL-divergence is 0, so that $q^{\backslash i}(f) \propto q(f)/\tilde{t}_i(f)$. This KL minimization can be easily done by the moment matching equations similar to (17) and (18):

$$\alpha^{\backslash i} = \alpha + \mathbf{h}^{\backslash i}\frac{\mathrm{d}\log\tilde{Z}_d}{\mathrm{d}m(\mathbf{x}_i)}, \quad \beta^{\backslash i} = \beta - \mathbf{h}^{\backslash i}\frac{\mathrm{d}^2\log\tilde{Z}_d}{(\mathrm{d}m(\mathbf{x}_i))^2}(\mathbf{h}^{\backslash i})^\mathrm{T} \tag{26}$$

where

$$\mathbf{h}^{\backslash i} \triangleq \mathbf{p}_i - \beta\boldsymbol{\phi}_i$$

$$\tilde{Z}_d = \int \frac{1}{\tilde{t}_i(f)}q^{\backslash i}(f)\mathrm{d}f$$

$$\propto \mathcal{N}(u_i|\mathbf{p}_i^\mathrm{T}\mathbf{m}_\mathbf{x}^{\backslash i}, -\tau_i + v(\mathbf{x}_i))$$

$$\frac{\mathrm{d}^2\log\tilde{Z}_d}{\mathrm{d}m(\mathbf{x}_i)^2} = \tau_i - v(\mathbf{x}_i)$$

$$\frac{\mathrm{d}\log\tilde{Z}_d}{\mathrm{d}m(\mathbf{x}_i)} = \frac{\mathrm{d}^2\log Z}{\mathrm{d}m(\mathbf{x}_i)^2}(m(\mathbf{x}_i) - g_i)$$

Note that we compute $m(\mathbf{x}_i) = \boldsymbol{\phi}_i^\mathrm{T}\boldsymbol{\alpha}$ and $v(\mathbf{x}_i) = \boldsymbol{\phi}_i^\mathrm{T}(\mathrm{diag}(\mathbf{w}) - \beta)\boldsymbol{\phi}_i = \boldsymbol{\phi}_i^\mathrm{T}\mathbf{h}^{\backslash i}$.

### 4.4 Eigenfunction selection

EigenGP uses a small subset of eigenfunctions. But unlike principle component analysis, we can improve the modeling power by *not* simply picking the top few eigenfunctions, as illustrated in Figure 2.(b) where the ninth, instead of the second, eigenfunction is selected. To select relevant eigenfunctions, we maximize the model marginal likelihood obtained from expectation progagation:

$$p(\mathbf{y}|\mathbf{w}, \mathbf{X}) = \int \mathcal{N}(\mathbf{f_x}|0, \tilde{\mathbf{K}}) \prod_i \tilde{t}_i(f)\mathrm{d}\mathbf{f_x}$$

$$= (\prod_i s_i)(\prod_j w_j^{\frac{1}{2}})|\mathbf{W} - \beta|^{\frac{1}{2}} \cdot$$

$$\cdot \exp(\frac{1}{2}(\alpha^\mathrm{T}(\mathbf{W} - \beta)^{-1}\alpha - \sum_i g^2(i)/\tau_i)) \tag{27}$$

We maximize the marginal likelihood via automatic relevance determination (ARD) (MacKay, 1992). To do so, we use an EP-EM approach. In the E-step, we compute $q(f)$ via EP. In the M-step, we maximize the expected likelihood $\mathbb{E}_q[\log p(\mathbf{y}, \mathbf{f_x}|\mathbf{w})]$ over $\mathbf{w}$, which leads to the following update:

$$w_j^{new} = w_j - \beta_{j,j} + \alpha_j^2. \tag{28}$$

Instead of the EM updates, we can use an active-set method (Faul & Tipping, 2002; Qi et al., 2004) to obtain efficient updates of $\mathbf{w}$. We can certainly explore other sparsification approaches, like $l_1$ penalty (i.e, a Laplace prior), an elastic net penalty, or spike and slab priors to sparsify $\mathbf{w}$. But a thorough comparison of various sparse priors is out of the scope of this paper.

### 4.5 Computational complexity

Since $\frac{\mathrm{d}^2 \log Z}{(\mathrm{d}m^{\backslash i}(\mathbf{x}_i))^2}$ is a scalar and $\mathbf{h}$ is a $L$ by 1 vector, it takes $O(L^2)$ to update $\beta$ via (18). Similarly, it takes $O(L^2)$ to update $\beta^{\backslash i}$ via (26). Therefore, given $N_l$ labeled training points, the computational cost is $O(L^2N_l)$ per EP iteration over all the data points. Since in practice the number of EP iterations is small (e.g., 10), the overall cost is $O(L^2N_l)$. Because of using the active-set method, the computation cost will be further significantly reduced to $O(r^2N_l)$ where $r$ is the actual number of eigenfunctions used in the EP iterations and $r < L$. In addition, we have the cost of computing the top $L$ eigenvectors $\mathbf{U}$ of the $Q$ by $Q$ kernel matrix $\mathbf{K_B}$; using efficient iterative algorithms such as the Lanzcos method, it takes $O(Q^2L)$. In summary, the complexity of the EP inference is $O(Q^2L + N_lL^2)$.

## 5 Related work

Our work is built upon the seminal work by Williams & Seeger (2001). But they differ in three critical aspects. First, we define a valid probabilistic model based on an eigen-decomposition of the GP prior. By contrast, the previous approach by Williams & Seeger (2001) aims at a low-rank approximation to the finite covariance/kernel matrix used in GP training – purely from a numerical approximation perspective – and its predictive distribution is not well-formed in a probabilistic framework (*e.g.*, it may give a negative variance of the predictive distribution). Accordingly, both predictive means and variances of these two methods are different. Second, while the previous Nyström method simply uses the first few eigenvectors, we maximize the model marginal likelihood to select eigenfunctions and adjust their weights, learning a nonstationary covariance function. Third, exploring the clustering property of the eigenfunctions of the Gaussian kernel, our approach can conduct semi-supervised learning, while the previous one cannot.

Our model also bears similarity to relevance vector machine (RVM) (Tippings, 2000) and sparse spectrum Gaussian process (SSGP) by Lázaro-Gredilla et al. (2010). But ours differs from these methods in four aspects. First, based on the K-L expansion of a covariance function, the basis functionsn of ours is quite different those used in RVM or SSGP. Second, with the white noise $\theta_0$ in the model, ours gives nonzero prediction uncertainty when a test sample is far from the training samples. In contrast, for this case the prediction of RVM shrinks to zero. Third, Figure 1 shows that our model has a two-layer structure, while RVM and SSGP do not. Note that while we estimate the parameters **u** and **w** (or equivalently $\theta$) associated with the two layers, our first layer training does not depend on the label **y** – in this sense, it echos the idea of unsupervised first layer training used in deep learning. Forth, our method can conduct semi-supervised learning while RVM and SSGP cannot.

## 6 Experiments

In this section we test EigenGP on regression, classification, and semi-supervised classification tasks.

### 6.1 Regression

For regression, we compared EigenGP with the full GP and two sparse approximate GP algorithms: the Nyström method (Williams & Seeger, 2001), and the pseudo-input approach, also known as Fully Independent Training Conditional (FITC) approximation (Snelson & Ghahramani, 2006). We used the Gaussian covariance (1) for all the competing methods. We optimized the kernel width $\eta$ for the full GP and used it for the Nyström method and EigenGP for a simple fair comparison. For the FITC, we used the optimization code from http://www.gatsby.ucl.ac.uk/~snelson/SPGP_dist.tgz to learn the kernel width for each dimension and the basis points. Because the optimization is sensitive to local optima (as observed by (Snelson & Ghahramani, 2006) and (Qi et al., 2010)), we tried different initial values to improve the results of FITC and presented here the best results we obtained (which are better than those of FITC using the kernel parameters learned from the full GP). For EigenGP, we set $w_0 = 0.1$ to have small white noise.

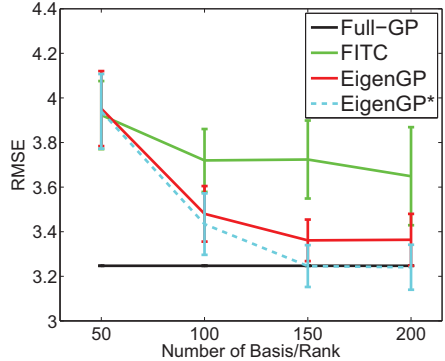

(a) Boston Housing

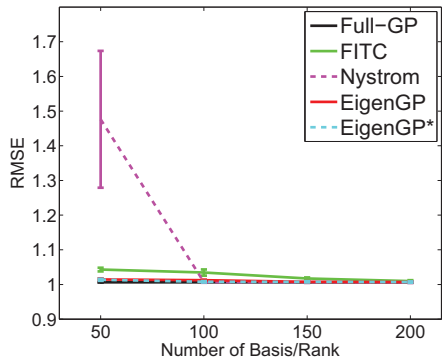

(b) Pumadyn-8nm

Figure 4: Regression results on Boston Housing and Pumadyn-8nm. The results of the Nyström method on Boston Housing are not included in the figure, since their errors are much larger than the others.

We compared the prediction accuracies of these alternative methods with the same computational complexity. Let $Q_1$ and $L$ be the number of basis points and the total number of eigenfunctions of EigenGP and the Nyström method, and $Q_2$ be the number of basis points of FITC. The computational complexities of EigenGP and the Nyström method is $O(Q_1^2 L + NL^2)$, while FITC takes $O(NQ_2^2)$. To make these algorithms have the same cost, we required $O(Q_1^2 L + NL^2) = O(NQ_2^2)$. To achieve this, we could set $L = Q_2$ and $Q_1$ is smaller than $\sqrt{NQ_2}$. In our experiments, we simply set $Q_1 = Q_2$ as well (This gave FITC more training time because of $N > Q_2$).

The results on two benchmark datasets, Boston Housing and Pumadyn-8nm, are reported in Figure 4. These two datasets contain 506 and 8192 data points. From these two datasets, we randomly selected 400 and 2000 points for training, respectively. And we used the rest for testing. We repeated these experiments 10 times and reported the average root mean square errors (RMSE) as well as the standard errors in Figure 4. For EigenGP, we not only ran the version with sparsification over **w** as described before, but also tested a version of EigenGP that simply chose the top $L$ eigenfunctions. We denote the later version as EigenGP*. The RMSE of the Nyström method on Boston housing are 312.5, 41.68, 15.17 and 6.480 with 50, 100, 150 and 200 basis points, respectively. The corresponding standard errors are 70.45, 11.23, 6.194 and 1.502. Despite the similarity between the two versions of EigenGPs and the previous Nyström method, EigenGPs significantly outperformed the Nyström method on Boston Housing consistently and on Pumadyn-8nm when the number of basis points is small the identical experimental setting – such as the same hyperparameters and the basis points between EigenGP and the Nyström method – in our experiments (because EigenGP* did not select eigenfunctions and estimate **w** via ARD, it even used the same weights as the Nyström method). This shows that when the number of basis points is small, the Nyström method suffers severely as the quality of the numerical approximation of the covariance matrix degenerates, while by contrast EigenGP and EigenGP*, as valid probabilistic models, degrade their performance smoothly.

### 6.2 Supervised classification

For supervised classification, we compared EigenGP with the full GP and three sparse GP algorithms: the Nyström method, Sparse Online Gaussian Processes (SOGP) (Csató & Opper, 2002), and FITC approximations (Snelson & Ghahramani, 2006; Naish-Guzman & Holden, 2008). We used the Gaussian covariance function for all these algorithms and tuned their kernel width by cross-validation. For the Nyström method, we used Laplace's method to approximate the posterior distribution as described in (Williams & Seeger, 2001). A better comparison with the Nyström method would be using EP, instead of the less effective Laplace's approximation. However, there is no previous work that combines EP with the Nyström method and the development of this algorithm is out of the scope of this paper. For the FITC model, we used EP for the posterior approximation as proposed by Naish-Guzman & Holden (2008). Although we did not maximize the model evidence to learn the kernel parameters for the FITC model (thus we may not obtain their best results), we found in practice that our extensive cross-validation of kernel parameters often gave prediction accuracies at least comparable to those based on evidence maximization.

We tested these algorithms on two benchmark datasets: Spambase and USPS. We randomly split the Spambase dataset into 2300 training and 2300 test samples 10 times and ran all the competing methods on each partition. For each partition, we chose the centers from K-means clustering of a randomly selected subset of the whole dataset as basis points for all the sparse GP methods. For the USPS dataset, we conducted three digit classification tasks: 8 vs 9, 3 vs 8, and 5 vs 8. Each digit in USPS was treated as a 256 dimensional vector. We randomly selected 1200 training and 1000 test samples and repeated the partition 10 times. In USPS, we used the same procedure to select basis points when the number of basis points is smaller than 500; when it is bigger than 500, we selected the basis points randomly.

As shown in Figure 5.(a), by sparsifying **w**, EigenGP consistently outperforms EigenGP* (which does not sparsify **w**) on the classification of digits 8 and 9. This improvement verifies the benefit of selecting relevant eigenfunctions for classification. Figures 5.(b)-(d) demonstrate that, with the same or less computational cost, EigenGP achieved lower classification error rates than the other sparse GP algorithms. Interestingly, EigenGP even outperforms the full GP. Two possible reasons are i) that by eigenfunctions selection, we obtain a nonstationary GP model that can better reflect the *local* smoothness of the latent function and remove labeling noise just like PCA for noise reduction, and ii) that the clustering property of eigenfunctions helps improve classification accuracy.

### 6.3 Semi-supervised classification

We compared EigenGP with three well-known semi-supervised learning algorithms: the graph regularization (GR) approach (Zhou et al., 2004), the Laplacian support vector machine (LAPSVM) (Belkin et al., 2006), and the sparse eigenfunction bases approach (SEB) (Sinha & Belkin, 2010). We did not include EigenGP* here because, without eigenfunction selection, it is not designed for semisupervised learning. For semi-supervised learning, we used both the labeled and unlabeled samples as basis points to generate the eigenfunctions for EigenGP. We also tested supervised SVM as a baseline for comparison. For all these al-gorithms, we used the Gaussian covariance function. The same kernel width was tuned by cross-validation.

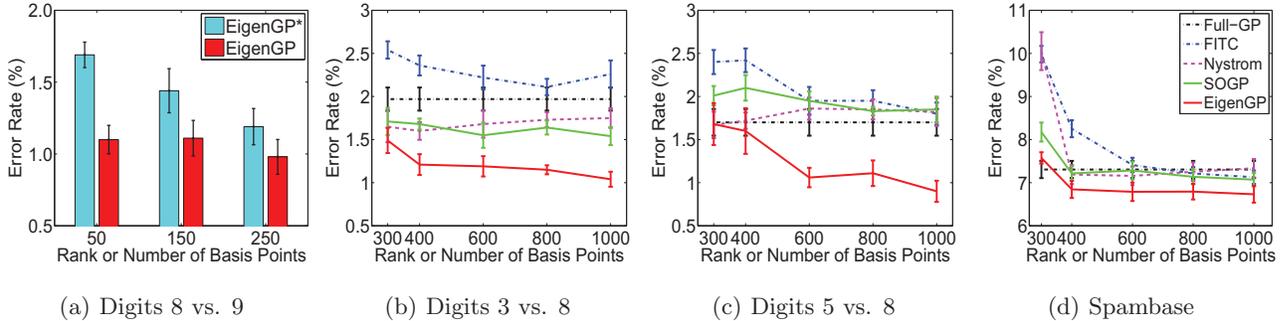

(a) Digits 8 vs. 9   (b) Digits 3 vs. 8   (c) Digits 5 vs. 8   (d) Spambase

Figure 5: Classification results on Spambase and USPS. The results are averaged on 10 random splits of the data and the error bars represent the standard errors.

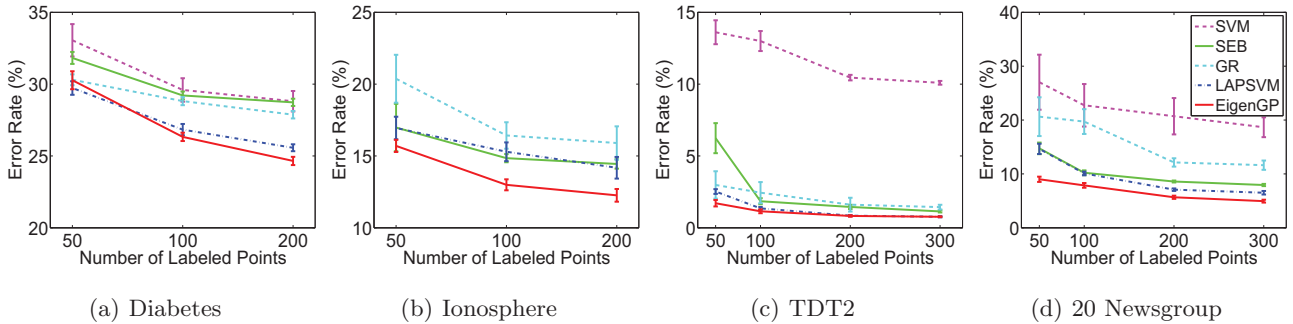

(a) Diabetes   (b) Ionosphere   (c) TDT2   (d) 20 Newsgroup

Figure 6: Semi-supervised classification results on Diabetes, Ionosphere, TDT2, and 20 Newsgroups. The results of SVM on Ionosphere are not reported here since they are much worse than the others.

We used two UCI datasets, Diabetes and Ionosphere, and two text datasets, TDT2 and 20 Newsgroups. Diabetes and Ionosphere contain 768 and 351 samples, respectively. For the TDT2 dataset, we selected two biggest categories for classification; for the 20 Newsgroup dataset, we chose two categories, comp.sys.ibm.pc.hardware and rec.sport.baseball, in our comparison. After the selection, we obtained 1976 and 3672 documents, respectively, from these two datasets. We then represented each document by the tf-idf term weights of the most frequent 1000 words.

We varied the number of randomly selected labeled samples and repeated the experiments 10 times. Figure 6 shows the averaged prediction error rates and the standard errors; clearly, EigenGP consistently outperformed the alternative approaches.

## 7 Conclusions

We have presented a new GP model, EigenGP, which selects eigenfunctions learned from data. Experiments demonstrated EigenGP's superior performance for regression, classification and semisupervised classification on several benchmark datasets. What is a little surprising is that, with lower computational cost, for classification tasks, it can outperform the full GP that uses infinite eigenfunctions. We believe the improvement in both training speed and prediction accuracy comes from the selection of relevant eigenfunctions and the nonstationarity in the covariance function associated with this selection.

We have used cross-validation to tune the kernel width of the covariance function. A future work is to learn the hyperparameters automatically from data. Although addressing this issue is out of the scope of this paper, we expect learning hyparameter for EigenGP is feasible by adopting a sampling method, for example, the slice sampling method proposed by Murray & Adams (2011).

Finally, we want to point out that the eigenfunctions in EigenGP can be viewed as dictionary elements. By sharing the dictionary elements across related tasks, EigenGP can be easily extended for multi-task learning.

## Acknowledgments

This work was supported by NSF IIS-0916443, NSF ECCS-0941043, NSF CAREER award IIS-1054903, and the Center for Science of Information, an NSF Science and Technology Center, under grant agreement CCF-0939370.